\documentclass[lettersize,journal]{IEEEtran}
\usepackage{amsmath,amsfonts} 
\usepackage{array}
\usepackage[caption=false,font=normalsize,labelfont=sf,textfont=sf]{subfig}
\usepackage{textcomp}
\usepackage{stfloats}
\usepackage{url}
\usepackage{verbatim}
\usepackage{graphicx}
\usepackage{multirow}
\usepackage{algorithm}
\usepackage{algpseudocode}
\usepackage{booktabs,array,xcolor}
\usepackage{cite}
\providecommand{\keywords}[1]{\textbf{\textit{Index terms---}} #1}

\begin{document}

\title{Architecture of an AI-Based Automated Course of Action Generation System for Military Operations}

\author{Ji-il Park,
        Inwook Shim*,
        Chong Hui Kim*
\thanks{Ji-il Park is with the Ministry of National Defense, 22 Itaewon-ro, Yongsan-gu, Seoul, Republic of Korea (e-mail: tinz64@kaist.ac.kr).}
\thanks{Inwook Shim is with the Department of Smart Mobility Engineering, Inha University, Incheon, Republic of Korea (e-mail: iwshim@inha.ac.kr).}
\thanks{Chong Hui Kim is with the Defense AI R\&D Institute, Agency for Defense Development (ADD), Daejeon, Republic of Korea (e-mail: chonghui.chkim@gmail.com).}
}

\markboth{}%
{Park \MakeLowercase{\textit{et al.}}: Architecture of an AI-Based Automated Course of Action Generation System}

\maketitle

\begin{abstract}
The automation system for Course of Action~(CoA) planning is an essential element in future warfare. As maneuver speeds increase, surveillance ranges extend, and weapon ranges grow, the operational area expands, making traditional manned-based CoA planning increasingly challenging. Consequently, the development of an AI-based automated CoA planning system is becoming increasingly necessary. Accordingly, several countries and defense organizations are actively developing AI-based CoA planning systems. However, due to security restrictions and limited public disclosure, the technical maturity of such systems remains difficult to assess. Furthermore, as these systems are military-related, their details are not publicly disclosed, making it difficult to accurately assess the current level of development. In response to this, this study aims to introduce relevant doctrines within the scope of publicly available information and present applicable AI technologies for each stage of the CoA planning process. Ultimately, it proposes an architecture for the development of an automated CoA planning system.

\end{abstract}

\keywords{
\small{CoA Automation, AI Staff Officer, Operational Execution Process, Decision Support.}
}

\maketitle

\section{Introduction}    
\label{sec:introduction}
As modern warfare evolves, the operational battlespace is expanding across multiple domains, including cyber, electromagnetic, land, air, and sea. The increasing complexity of military operations—driven by accelerated maneuver speeds, extended surveillance capabilities, and advanced weaponry—has significantly challenged traditional human-centric Course of Action~(CoA) planning~\cite{b1,b2}. Commanders and staff must now process vast amounts of information from diverse sources while considering an increasing number of operational factors under conditions of heightened uncertainty.

To cope with these challenges, advancements in artificial intelligence (AI) have led to the active development of AI-based intelligent command decision support systems. These systems are designed to support commanders by enhancing decision-making speed and accuracy, particularly in high-tempo, multi-domain operations \cite{b3,b4,b5}. The growing reliance on real-time data from sensors across space, cyber, and electromagnetic domains further increases the need for automation in the operation process \cite{b6}. Additionally, reductions in military personnel and subsequent force structure transformations—such as the integration of manned–unmanned teaming systems—have expanded operational areas, reinforcing the necessity for AI-driven solutions \cite{b7}. In this context, intelligent command decision support systems play a crucial role in mitigating battlefield uncertainty and enabling rapid and precise decision-making. Among these, CoA automation systems represent a core technological capability that underpins effective AI-driven operational planning \cite{b8,b9}. Recent doctrinal and capability development efforts by major defense organizations further emphasize the institutionalization of AI-enabled decision support tools within multi-domain operational planning frameworks \cite{b2,b9}.

To address these challenges, many nations and defense companies are developing AI-based CoA planning systems to enhance decision-making and operational effectiveness. However, due to strict security requirements and proprietary constraints in military weapon system development, publicly available technical details and empirical evaluations of end-to-end CoA automation systems remain limited.

To this end, this study aims to propose an architecture for the development of an AI-based automated CoA system by considering publicly available doctrinal information and the current technological level \cite{b1}. Through an examination of publicly available doctrines, we establish a foundational understanding of CoA development and its role across cyber, electromagnetic, land, air, and sea domains. Additionally, we explore AI technologies applicable to each stage of the CoA planning process, including data fusion, threat analysis, predictive modeling, and adaptive decision-making \cite{b6}. Based on these findings, we propose a conceptual framework for an AI-powered CoA planning system capable of operating across multiple domains.

Our approach emphasizes the development of multi-modals for learning and processing different forms of data such as photography, video, speech, text, and supervised data, along with the integration of artificial intelligence techniques such as machine learning, deep learning, and reinforcement learning \cite{b3}, to enhance situational awareness and improve operational agility. By identifying key areas for future research of AI-based CoA systems, the proposed architecture accelerates the development of AI-based decision support systems, enabling more intelligent, adaptive, and efficient military operation planning in increasingly complex battlespaces.

\section{Related Works}
Decision-support systems for military operational planning have been extensively studied to assist commanders in understanding battlefield situations and improving operational effectiveness. Existing approaches can generally be categorized into three major paradigms: rule-based systems, end-to-end learning models, and AI-driven analytical frameworks. This section reviews representative studies in each category, highlighting their characteristics and limitations while motivating the integrated AI architecture proposed in this study.

\subsection{Rule-Based Models}
Traditional military decision-support systems often rely on rule-based reasoning frameworks constructed from explicitly defined doctrinal rules and expert knowledge. These systems offer high interpretability and reliability, which are essential in military contexts where transparency and doctrinal consistency are required.

Knowledge requirements and management strategies for expert systems in military situation assessment have been investigated in earlier studies \cite{b35}. In this work, knowledge structures were divided into \emph{global knowledge}, responsible for identifying relevant hypotheses, and \emph{local knowledge}, used to evaluate specific alternatives. Case-based decision support systems (CBDSS) have also been introduced for command and control training environments \cite{b36}. By combining case-based reasoning with decision-support mechanisms, such systems assist military personnel in learning and applying standard operating procedures (SOPs).

Another line of research has focused on the validation of complex military simulations using knowledge-based approaches \cite{b37}. These methods reduce the reliance on extensive expert evaluation by embedding domain knowledge into simulation validation processes. Multi-agent-based simulation frameworks have also been developed to represent land combat operations as complex adaptive systems \cite{b38}. Such frameworks demonstrate the potential of agent-based modeling as a decision-support tool for operational analysis.

In addition, agent-based modeling techniques have been applied to simulate small-unit military operations \cite{b39}. These models explicitly incorporate human-related factors such as leadership, morale, and training level, all of which significantly influence combat effectiveness and operational outcomes.

\subsection{End-to-End Models}
Unlike rule-based approaches, end-to-end models attempt to learn direct mappings from raw input data to operational outputs without relying on manually designed intermediate representations. These models are typically built upon deep learning techniques, including convolutional neural networks (CNNs) and deep reinforcement learning (DRL).

Recent studies have explored end-to-end frameworks for resilient connectivity in UAV communication networks operating under adversarial jamming environments \cite{b40}. Signal-strength prediction models have also been developed for underwater optical communication systems to improve relative pose estimation and modem performance in complex environments \cite{b41}. 

From a broader system integration perspective, modeling, simulation, and visualization frameworks have been proposed to combine terrestrial, airborne, and space environments for comprehensive mission analysis \cite{b42}. Research has also addressed interoperability challenges across heterogeneous security domains by proposing architectures that enable unified service management across multiple military network infrastructures \cite{b43}.

Despite their advantages in scalability and real-time processing capability, end-to-end learning models are often criticized for their limited interpretability. In military decision-making environments—where accountability and doctrinal justification are required—this lack of explainability remains a significant concern.

\subsection{AI-Based Models}
Recent advances in artificial intelligence have led to the development of AI-based methodologies that utilize supervised learning, unsupervised learning, deep neural networks, and reinforcement learning to support complex military decision-making processes. These approaches aim to enhance adaptability, predictive capability, and real-time responsiveness in dynamic operational environments.

Comprehensive reviews have discussed the strategic implications of artificial intelligence across various military domains \cite{b44}. Defense initiatives in the United States and Europe have also been analyzed to illustrate ongoing efforts to integrate AI technologies into military systems and enhance operational readiness \cite{b45}. In addition, the expanding role of AI in logistics, object detection, and autonomous platforms has been examined, together with concerns regarding potential geopolitical instability and escalation risks \cite{b46}. Artificial intelligence has also been characterized as a “technology multiplier” that accelerates tactical innovation and highlights the importance of improved simulation environments and knowledge engineering techniques \cite{b47}.

In addition, recent studies have investigated AI-based techniques for automated Course of Action (CoA) analysis in military wargaming environments. AI-enabled wargaming platforms have been proposed to support CoA evaluation within the Military Decision Making Process (MDMP) \cite{b47a}. More recently, large language models have shown potential for rapid CoA generation, highlighting the emerging role of generative AI in military operational planning \cite{b47b}.

Recent studies have further explored the strategic reasoning capabilities of frontier large language models in simulated military crisis scenarios \cite{b47c}. In these studies, advanced language models were evaluated in adversarial simulations in which the models assumed the roles of opposing national leaders during crisis decision-making processes. The results demonstrate that modern large language models are capable of interacting within simulated adversarial environments, generating strategic responses, and evaluating escalation dynamics under uncertainty.

In addition to simulation-based studies, recent reports indicate that advanced AI systems have begun supporting real-world military operations. For example, Anthropic’s large language model Claude has been integrated into the U.S. military’s intelligence analysis platform, the Maven Smart System, where it assists in analyzing intelligence data and prioritizing potential targets during military operations \cite{b47d}. These developments suggest that large language models may play an increasingly important role in future AI-driven wargaming environments and automated military decision-support systems.

Recent studies have also explored the strategic reasoning capability of frontier large language models in simulated military crisis scenarios. In particular, models such as Claude, GPT, and Gemini have been used to analyze contemporary conflict situations, including scenario-based assessments related to the Iran–Israel conflict. In these experiments, the models generated and evaluated alternative attack and response strategies through iterative scenario simulations, enabling the analysis of escalation risks and damage minimization strategies. The results indicate that modern large language models have advanced to a level where they can interact within simulated adversarial environments and perform strategic reasoning across complex conflict situations. These developments suggest that large language models may play an important role in future AI-driven wargaming environments and automated military decision-support systems \cite{b47c}.

Recent studies have also examined the strategic risks associated with AI-enabled military competition, proposing confidence-building
measures (CBMs) to improve transparency and reduce instability \cite{b48}. In addition, AI-driven wargaming environments have been explored to support operational planning and decision analysis \cite{b47a}. In addition, the convergence of artificial intelligence with cyber warfare has been examined, demonstrating how AI technologies are reshaping modern digital battlefields \cite{b50}.

Despite the advantages of AI-based approaches in terms of adaptability and situational awareness, several challenges remain. Issues related to explainability, security, and robustness continue to limit their deployment in mission-critical military applications. Furthermore, although considerable progress has been achieved across rule-based, end-to-end, and AI-driven approaches, existing operational planning systems remain fragmented and domain-specific, thereby restricting comprehensive situational awareness and
hindering consistent decision-making.

However, most existing studies primarily focus on AI technologies, simulation environments, or automated wargaming platforms, without systematically analyzing the doctrinal structure and step-by-step procedures of military operational planning. In particular, limited research has investigated how AI systems can be integrated with doctrine-based operational planning processes such as the Military Decision Making Process (MDMP) or Intelligence Preparation of the Battlefield (IPB). Consequently, the development of an AI architecture that explicitly incorporates doctrinal planning procedures and supports the entire operational planning cycle remains largely unexplored in existing research. This limitation motivates the need for an integrated framework
that combines doctrinal reasoning with modern AI technologies. 

To address these limitations, this study proposes an integrated AI architecture that spans the entire military operational lifecycle—from \textit{Intelligence Preparation of the Battlefield (IPB)} to mission execution and post-operation evaluation. The proposed framework integrates the interpretability of rule-based models, the computational efficiency of end-to-end learning approaches, and the adaptability of modern AI techniques. Through this integration, the architecture aims to support robust and transparent decision-making across multi-domain military operations while maintaining continuous situational awareness throughout the operational process. The main contributions of this work are summarized as follows:

\begin{enumerate}
\item An integrated multi-paradigm decision-support architecture that combines rule-based reasoning, end-to-end learning, and AI-based analytical models for consistent and reliable military operational planning.

\item A unified architecture covering the entire operational lifecycle—from Intelligence Preparation of the Battlefield (IPB) to mission execution and post-mission evaluation—thereby enabling continuous situational awareness and seamless decision support across all operational phases.
\end{enumerate}

\section{prerequisites}  
\subsection{Operation Process}  

Course of Action (CoA) development is a critical phase within the operation process; therefore, a comprehensive understanding of the overall operation process is essential for developing an automated CoA planning system. Accordingly, the discussion on CoA development will be conducted within the broader context of the operation process. To achieve this, an overview of the operation process will first be provided, followed by an in-depth examination in the main section of applicable artificial intelligence (AI) technologies and their implementation strategies for each phase.

The operation process refers to a structured sequence of activities required to conduct military operations and consists of four key phases: Planning, Preparation, Execution, and Assessment. Commanders and staff oversee and direct operations throughout these phases to ensure mission accomplishment and operational objectives are met. This study focuses on proposing an architecture for the development of the Planning phase \cite{b59}. 

\begin{figure}[t!] 
\centering
\includegraphics[width=1.0\linewidth]{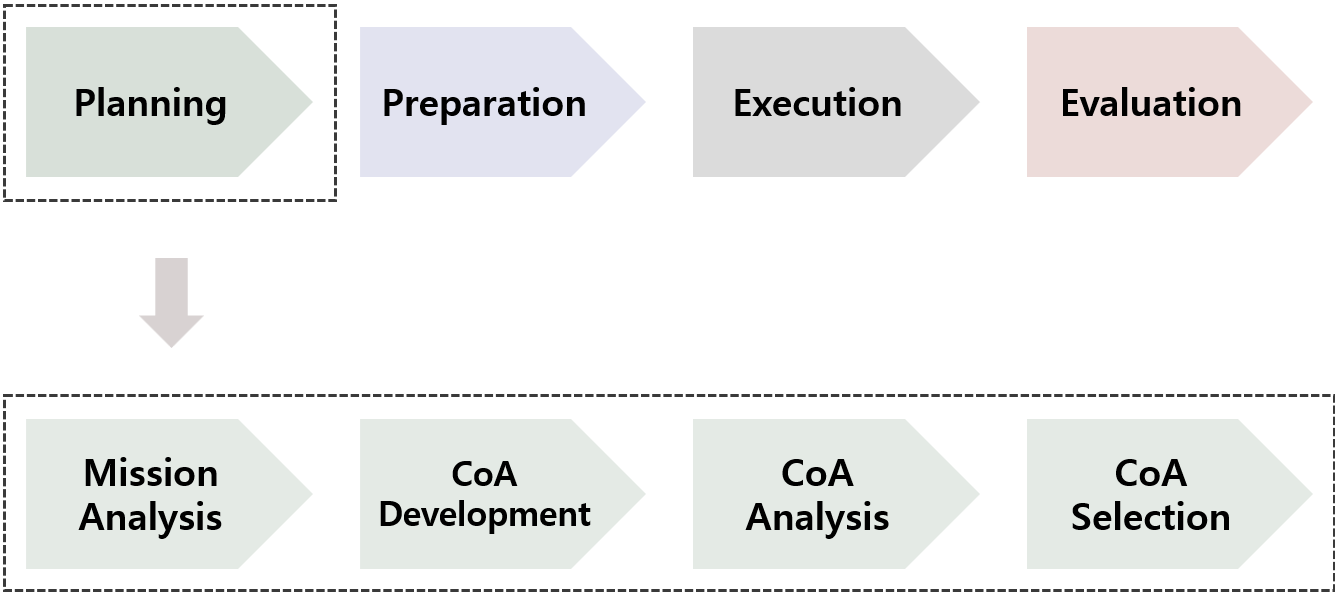}
\caption{Overview of the planning phase in the operational execution process \cite{b59}.}
\label{fig:mission}
\end{figure}

The Planning phase comprises four fundamental components: mission analysis, CoA development, CoA analysis, and CoA decision. Mission analysis assesses the operational environment, objectives, constraints, and resources to define mission requirements. CoA development formulates feasible options aligned with mission objectives, considering factors like enemy capabilities, terrain, and assets. CoA analysis evaluates these options through wargaming, simulations, and risk assessments. Finally, CoA decision selects the most effective course of action to achieve mission success while minimizing risks. The development of an automated Course of Action (CoA) establishment system primarily refers to the automation of the planning phase. Therefore, as illustrated in Fig. \ref{fig:mission}, this paper focuses on the planning phase.

While the operation process is generally conducted sequentially, certain operational conditions, particularly those with time constraints, may necessitate the concurrent execution of multiple phases. For reference, while each phase is typically conducted in a sequential manner, there are situations where time constraints necessitate their simultaneous execution.

\subsection{Intelligence Preparation of the Battlefield}
Intelligence Preparation of the Battlefield (IPB) is a critical component in the development of an automated Course of Action (CoA) planning system, as it is continuously conducted throughout the phases of planning, operational preparation, and execution. The IPB process consists of four key stages: battlespace evaluation, battlespace analysis, enemy capability assessment, and enemy CoA analysis \cite{b60}. Through these stages, the operational environment is systematically assessed, leading to the development of an Operational Area Analysis Map. Based on this assessment, an Enemy Situational Map is generated, followed by the identification  of the most feasible enemy CoAs. Since the formulation of operational plans is heavily dependent on the anticipated enemy CoAs, this process plays a decisive role in ensuring strategic and tactical effectiveness.

If the IPB process is not conducted with precision, the accuracy of enemy CoA predictions is significantly diminished, potentially resulting in erroneous strategic decisions. Consequently, maneuver plans derived from inaccurate enemy CoA assessments may fail to achieve their intended operational objectives, leading to compromised mission success and increased risks to forces on the ground. Moreover, as modern warfare increasingly incorporates multi-domain operations (MDO), the complexity of battlespace analysis has grown, necessitating advanced analytical tools, such as artificial intelligence (AI) and machine learning (ML), to enhance IPB accuracy. By integrating AI-driven predictive analytics and real-time data fusion, military planners can improve the reliability of enemy CoA forecasts and optimize decision-making processes.

In addition, the automation of IPB can contribute to the development of decision support systems (DSS) that enhance commanders' ability to process vast amounts of intelligence more efficiently. This is particularly significant in high-tempo operational environments where rapid and informed decision-making is essential. By leveraging AI-based IPB methodologies, military forces can achieve improved situational awareness, reduce cognitive load on decision-makers, and enhance the overall agility of combat operations. Therefore, ensuring the accuracy and reliability of the IPB process is paramount in modern military operations, particularly in the automation of CoA planning systems.

\begin{figure}[t!] 
\centering
\includegraphics[width=1.0\linewidth]{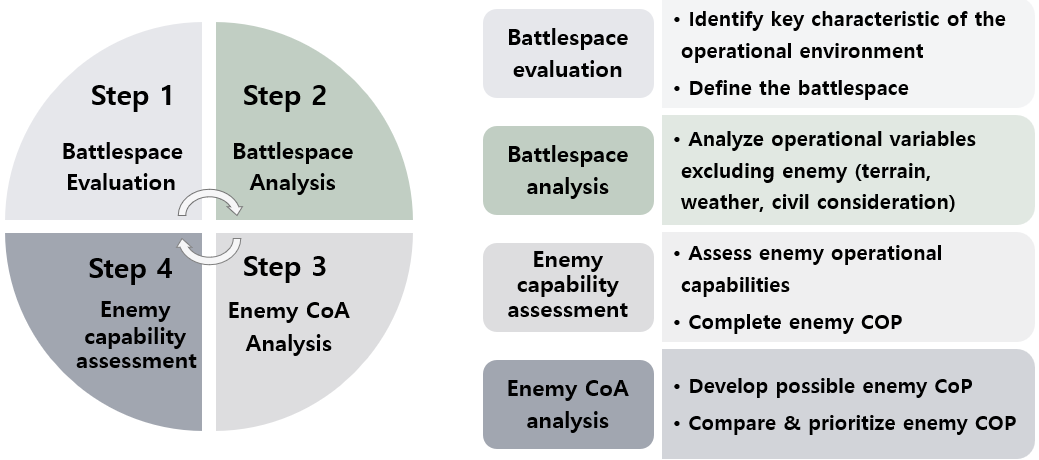}
\caption{Intelligence Preparation of the Battlefield (IPB) Process and Major Outputs at Each Step \cite{b60}.}
\label{fig:step}
\end{figure}

\begin{figure}[b!] 
\centering
\includegraphics[width=1.0\linewidth]{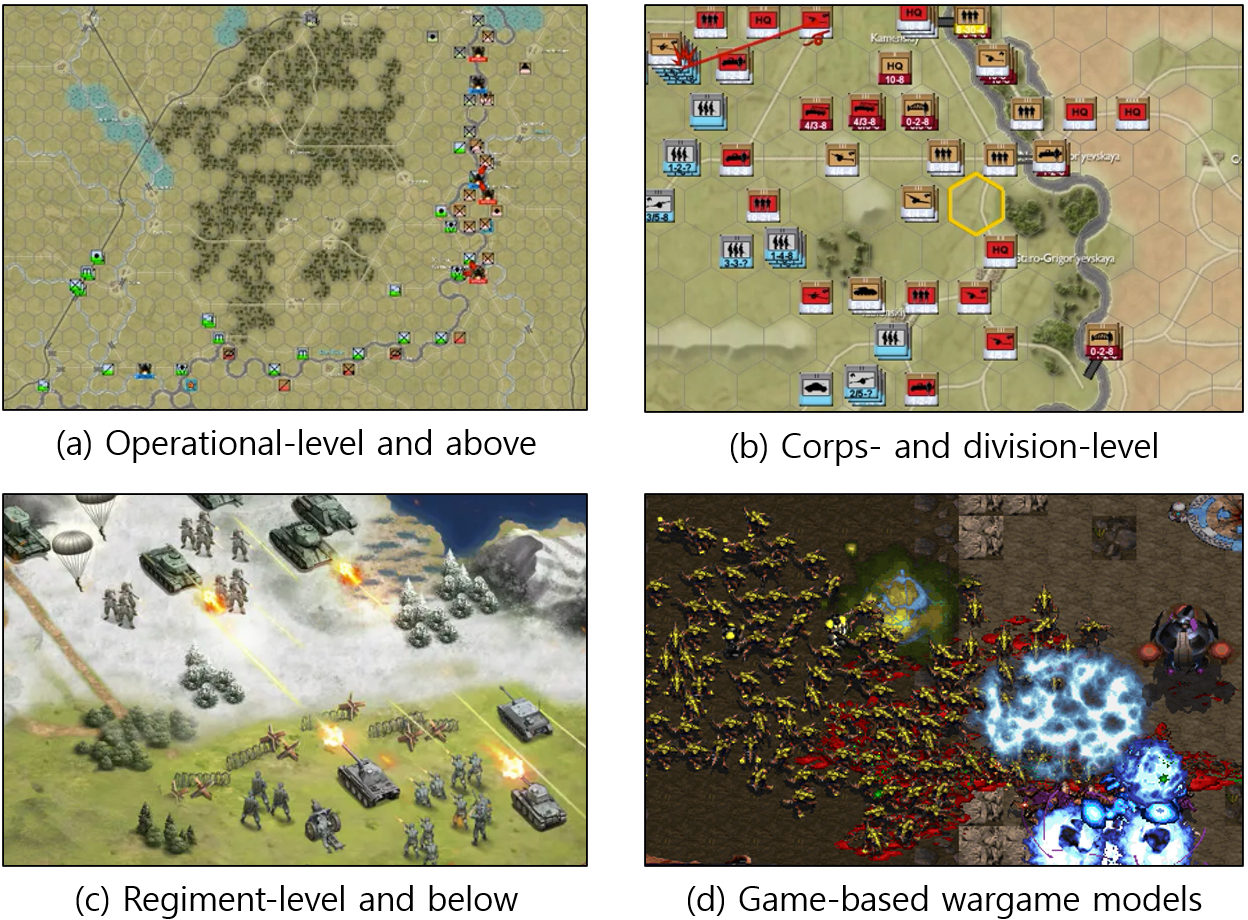}
\caption{Concept of differential implementation of echelon-specific wargaming models for unit-level course of action development.}
\label{fig:war1}
\end{figure}

\subsection{Function of Combat Power}

The Command and Control (C2) function enables commanders and staff to direct and manage subordinate units, serving as the most critical component among the six warfighting functions. It plays a leading role in integrating all available operational elements to ensure effective mission execution. Commanders receive a wide range of information related to the other warfighting functions—Intelligence, Movement and Maneuver, Fires, Protection, and Sustainment—which allows them to make timely and informed decisions before issuing orders to subordinate units.

The Intelligence function provides the necessary intelligence and information to support mission execution while ensuring information superiority for efficient employment of combat power. Its primary roles include minimizing battlefield uncertainty, enhancing battlefield visualization, and denying the enemy access to critical information. It delivers real-time battlefield intelligence, including enemy disposition, terrain, and weather conditions, through Intelligence Preparation of the Battlefield (IPB) analysis to assist commanders in making tactical and strategic decisions.

\begin{figure}[b!] 
\centering
\includegraphics[width=1.0\linewidth]{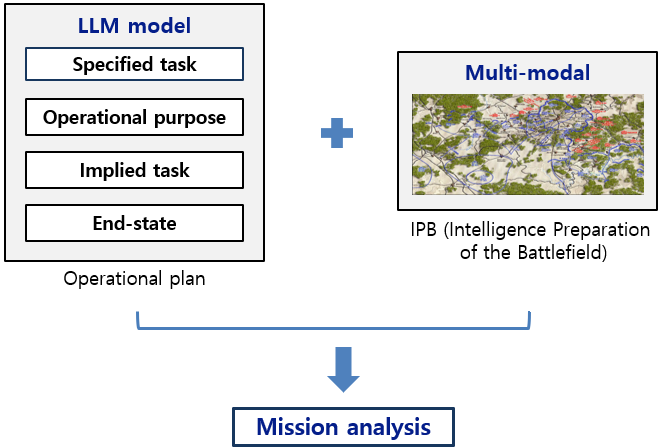}
\caption{Mission Analysis via LLM-Based Operational Plan Interpretation and Multi-Modal IPB.}
\label{fig:process}
\end{figure}

The Movement and Maneuver function focuses on positioning and deploying forces to establish favorable conditions for friendly troops while striking the enemy’s center of gravity and vulnerabilities through surprise attacks, thereby disrupting the enemy’s structure. While it may appear similar to C2, the key distinction lies in their focus: C2 involves comprehensive decision-making considering all warfighting functions, while Movement and Maneuver focuses primarily on mobility and tactical execution with partial consideration of intelligence and fires functions.

\begin{figure*}[t!] 
\centering
\includegraphics[width=0.995\linewidth]{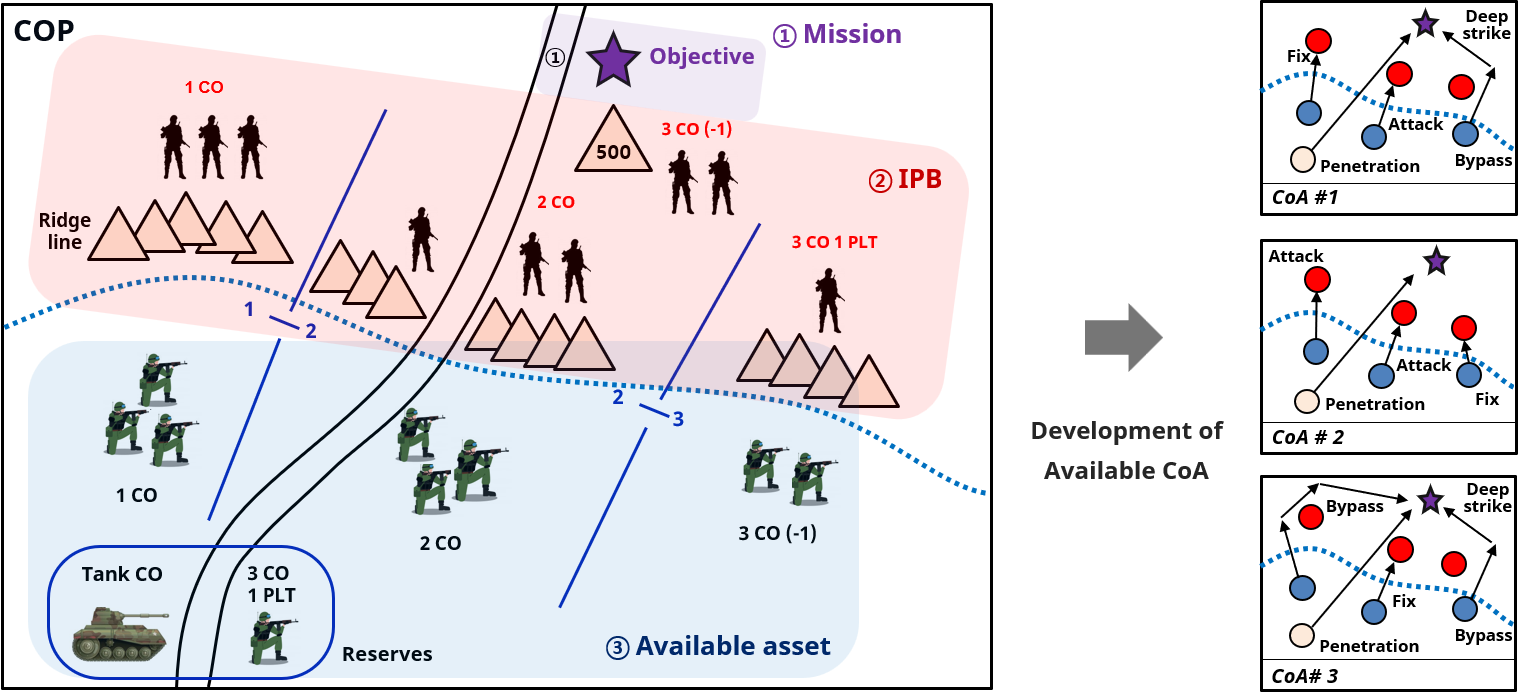}
\caption{Generation of multiple candidate courses of action using artificial intelligence.}
\label{fig:CoAs}
\end{figure*}

\begin{table*}[h!]
\centering
\renewcommand{\arraystretch}{1.75}
\setlength{\tabcolsep}{6pt}
\begin{tabular}{|p{3.0cm}|p{13.75cm}|}
\hline
\textbf{Section} & \textbf{Description} \\
\hline
\textbf{1. Situation} & 
\textbf{a. Enemy Forces}: Enemy disposition, location, capabilities, and activities \newline
\textbf{b. Friendly Forces}: Higher headquarters’ mission and intent; status of adjacent and supporting units \newline
\textbf{c. Attachments \& Detachments} \\
\hline
\textbf{2. Mission} &
Clear mission statement including the 5W (Who, When, Where, What, Why). \newline
\emph{Example:} “The 1st Infantry Battalion seizes Objective XYZ at 0600 on 20 March 2025 in order to prevent enemy advance.” \\
\hline
\textbf{3. Execution} &
\textbf{a. Commander’s Intent}: End state, key tasks, and success criteria \newline
\textbf{b. Concept of Operations}: Operational approach and main phases \newline
\textbf{c. Tasks to Subordinate Units} \newline
\textbf{d. Coordination \& Control}: Required coordination during the operation \\
\hline
\textbf{4. Sustainment} &
\textbf{a. Logistics}: Supply (fuel, ammunition, rations) \newline
\textbf{b. Medical Support}: Medical unit deployment and evacuation plan \newline
\textbf{c. Transportation}: Movement methods and main routes \\
\hline
\textbf{5. Command \& Signal} &
\textbf{a. Command}: Command relationships and command post locations \newline
\textbf{b. Signal}: Frequencies, passwords, call signs \\
\hline
\end{tabular}
\caption{Example of mission analysis for illustrative purposes.}
\label{tab:hhq_oplan}
\end{table*}

The Fires function pertains to the projection of firepower against enemy targets, determining the volume of fire and strike capabilities. Since targets continuously move in real-time during operations, commanders must rapidly select available artillery units and execute swift target engagement procedures. If an AI-driven automated target processing system is integrated into the Fires function, it can significantly enhance response times and the effectiveness of fire support.

Protection aims to preserve friendly combat power, encompassing measures such as Chemical, Biological, Radiological, and Nuclear (CBRN) protection and base defense. Among the five major enemy threats, CBRN attacks can be anticipated through AI-based battlefield information analysis. Predictive capabilities allow proactive preparation of decontamination facilities in contaminated areas, thereby minimizing potential damage.

Finally, the Sustainment function ensures operational endurance by managing and providing essential resources to maintain combat effectiveness. In logistics, AI applications can enhance real-time ammunition and supply distribution, while in personnel management, AI can optimize troop control and reinforcement, ensuring the continuous support of operational units.


\subsection{Operational Art and Tactics}
\subsubsection{Operational Art}
The levels of war are categorized into tactics, operational art, and strategy. Generally, tactics refer to the employment of forces at the corps level and below, operational art pertains to the orchestration of forces at the operational echelon, and strategy concerns the employment of forces at the theater and national levels. While these concepts can be somewhat complex, they can be better understood through a quote from Alexander Svechin’s book Strategy, which states, “'Tactics make the steps from which operational leaps are assembled”
This analogy helps clarify the distinction between the different levels of warfare. Among these, strategy involves not only military aspects but also various factors such as diplomacy and politics. Therefore, this research will focus solely on operational art and tactics, excluding strategy.

\begin{figure*}[t!] 
\centering
\includegraphics[width=0.995\linewidth]{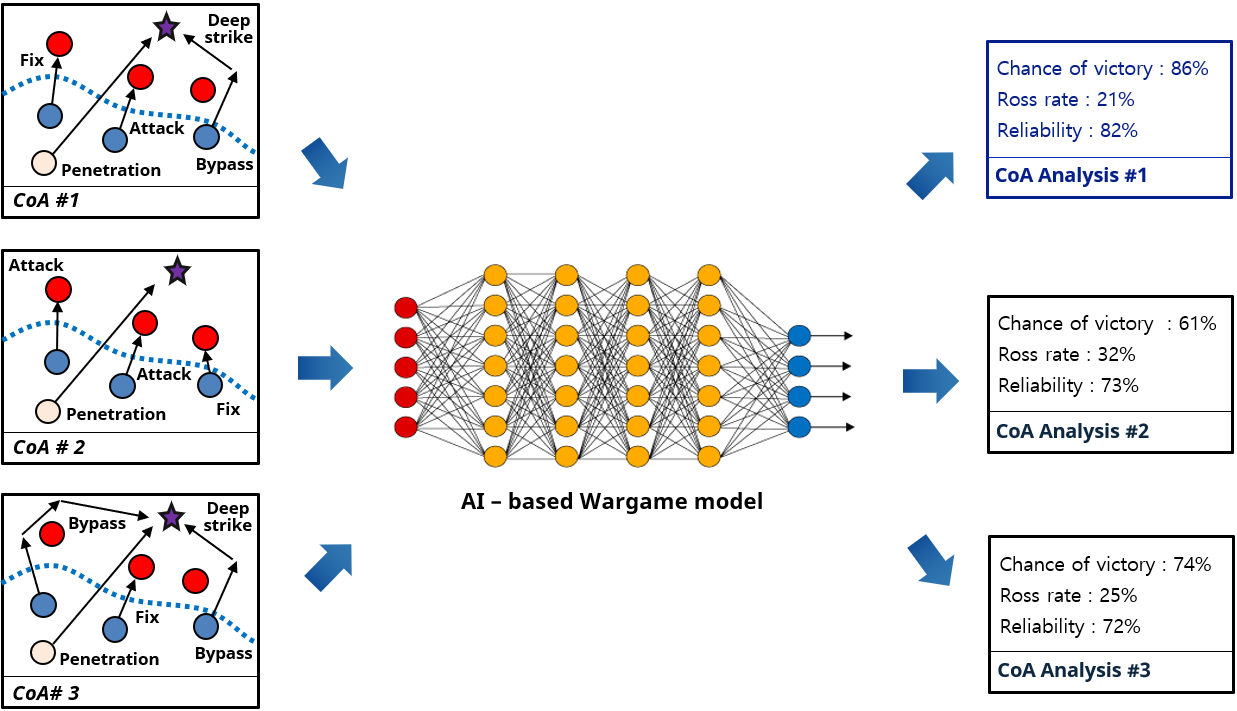}
\caption{AI-Based Wargaming Simulation for Evaluating Multiple Courses of Action Using Quantitative Metrics.}
\label{fig:wargame}
\end{figure*}

\begin{table*}[h]
\centering
\renewcommand{\arraystretch}{1.65}   
\begin{tabular}{|p{2.5cm}|p{7.35cm}|p{6.5cm}|}
\hline
\textbf{Detailed Step} & \textbf{Description} & \textbf{Result (Example)} \\ \hline
Specified task &
Confirmation of the clearly assigned task issued by the higher headquarters &
Secure Objective Area 00 \\ \hline
Operation purpose &
Identification of the primary purpose of the operation &
Ensure maneuver conditions for follow-on forces \\ \hline
Implied task &
Identification of additional tasks determined by the commander to accomplish the specified task &
Secure Route 00 to enable seizure of Objective Area 00 \\ \hline
Constraint &
Identification of anticipated constraints during the operation &
River flowing from east to west \\ \hline
End-state &
Description of the desired conditions at the conclusion of the operation &
Enemy neutralized within the operational area \\ \hline
Mission statement &
Selection of tasks to accomplish the operational purpose &
-- \\ \hline
\textbf{Mission} &
\multicolumn{2}{p{11cm}|}{\textbf{Secure Objective 00 and ensure maneuver conditions for follow-on forces}} \\ \hline
\end{tabular}
\caption{Example of Mission Analysis for Course of Action (CoA) Development.}
\end{table*}

First, the development of an automated CoA (Course of Action) planning system at the operational level should incorporate Modeling and Simulation (M\&S) methodologies. This is because operational-level formations cover vast areas of operation, involve numerous units and personnel, and utilize diverse weapon systems, making it impractical to depict every entity at the individual level. For example, as shown in figure, terrain is often simplified into hexagonal or square grid-based maps, and instead of representing individual soldiers or weapon systems as distinct entities, units are displayed using military symbols. Each unit is then assigned a combat power index that takes into account the characteristics of its weapon systems and military branches, allowing simulations to be conducted efficiently.

In other words, at the operational level, military operations do not depict the detailed tactics of subordinate units; rather, combat outcomes are represented using simplified combat power indices. Additionally, operational objectives differ from those at the tactical level—whereas tactical echelons focus on securing key terrain, operational-level objectives target the "core" of enemy forces, such as an enemy group army command post. These distinctions must be considered in the development of an automated CoA planning system.

\subsubsection{Tactics}
On the other hand, tactical echelons below the corps level should describe operations in a more detailed manner compared to operational echelons. First, since corps- and division-level units have relatively large operational areas and possess numerous subordinate units, it is effective to depict objects at the battalion and company levels, as illustrated in <Figure> (a). Meanwhile, at the battalion and company levels, where the operational area is smaller and the variety of forces and weapon systems is limited, it is possible to represent each object and weapon system individually, as shown in <Figure> (b).


One important factor to consider when developing a wargame model for the establishment of tactical echelon-level operational plans is that the operational area (battle boundary) of each echelon must be clearly defined, ensuring that objects operate only within their assigned operational zones.


For example, when using a game to develop a model, it would be acceptable if all 100 attacking units belonged to the same company. However, if units from different companies were mixed together, the simulation would fail to accurately represent real combat situations. Additionally, since there are leadership units such as platoon leaders among them, they should be grouped into platoon-sized units for movement to ensure a realistic combat simulation. Therefore, it is crucial to take these factors into account when developing a wargame model to enhance the accuracy and realism of tactical echelon-level planning systems.

\section{AI Technology Application Strategies for Each Phase} 
As previously mentioned, the development of the CoA automation system refers to the automation of the planning phase. Therefore, this chapter focuses on the planning phase, which is the first step of the operation execution process, and addresses in detail the IPB procedures that are continuously conducted throughout the entire operation execution process.

\subsection{Operational Execution Process: Planning}

\subsubsection{Mission analysis}
Mission analysis is the process of analyzing a unit’s mission by identifying the higher commander’s intent and specified tasks, followed by the derivation of the unit’s operational objectives and inferred tasks. The higher commander’s intent refers to the intent of both the first and second higher-level commanders (e.g., for a brigade commander, this would include the division and corps commanders). Specified tasks are those explicitly assigned by the higher echelon. As shown in Table \ref{tab:hhq_oplan}, both the higher commander’s intent and specified tasks are typically included in the operation order issued by the higher unit and can therefore be extracted through text analysis. The operational objective refers to the purpose of the operation to be carried out by the unit, while the inferred tasks are additional tasks identified by the commander based on their own judgment, deemed necessary to achieve the higher commander’s intent, fulfill the specified tasks, and accomplish the operational objective.

As shown in Table \ref{tab:hhq_oplan}, an operational order is composed of five main components—Situation, Mission, Execution, Sustainment, and Command and Control—each of which contains several detailed subcomponents. Since the operational order is entirely written in text-based format, it is possible to utilize an LLM model trained on such data to extract specified tasks, operational objectives, and inferred tasks, as illustrated in Table \ref{tab:hhq_oplan}, thereby enabling comprehensive mission analysis.

However, mission analysis should not rely solely on the text-based operational order; it must be continuously updated by incorporating real-time battlefield information gathered throughout the entire operational execution process. Therefore, as shown in Fig. \ref{fig:process}, the results of a multimodal IPB (Intelligence Preparation of the Battlefield) analysis model should also be integrated to derive a comprehensive and accurate mission analysis. Inaccurate mission analysis significantly increases the likelihood of formulating flawed courses of action. Therefore, during the development of an automated course-of-action planning system, this phase demands the highest level of precision among all stages of the process.

\begin{figure}[b!] 
\centering
\includegraphics[width=1.0\linewidth]{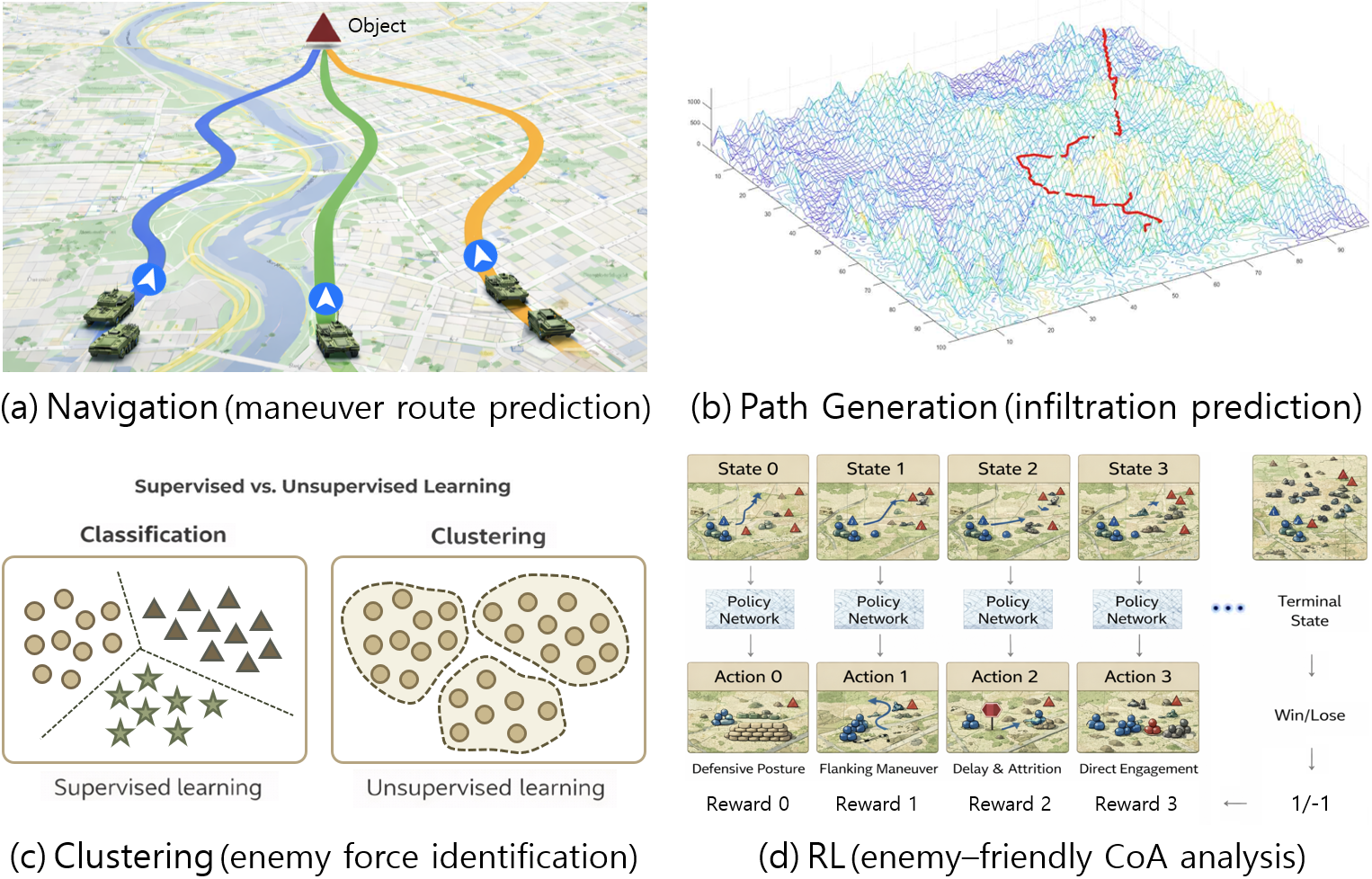}
\caption{Representative AI Techniques Applicable to Enemy Infiltration Analysis \cite{b55}, Force Identification, and CoA Development.} 
\label{fig:app_sample}
\end{figure}

\subsubsection{CoA development}
Due to the complexity of concretizing courses of action, the process is illustrated through a simplified example, as shown in Fig. \ref{fig:CoAs}, to aid understanding. The example represents a battalion-level operation, which is significantly less complex than actual scenarios; in reality, the complexity increases substantially with the size of the unit.

As illustrated in Fig. \ref{fig:CoAs}, the mission analysis concluded that the objective is to secure Hill 500 and Route 1, as marked within the purple box. The battlefield information analysis indicates that the terrain is characterized by laterally developed ridge lines, which provide a defensive advantage to enemy forces. In contrast, Route 1 extends longitudinally, thereby facilitating maneuverability for friendly forces. Enemy forces are estimated to comprise two companies and one platoon deployed in the forward area, with one company (-) positioned in the rear. These analytical results are highlighted in the red box.

The blue box indicates the assets and capabilities available to the friendly forces. Based on this analysis, multiple feasible courses of action can be derived. Provided that data on viable infiltration and maneuver routes, enemy dispositions, and friendly assets and capabilities are available, current AI technologies are sufficiently advanced to generate all possible CoA options. The figure also illustrates key AI component technologies that can be utilized in deriving these feasible plans.

\begin{figure}[t!] 
\centering
\includegraphics[width=1.0\linewidth]{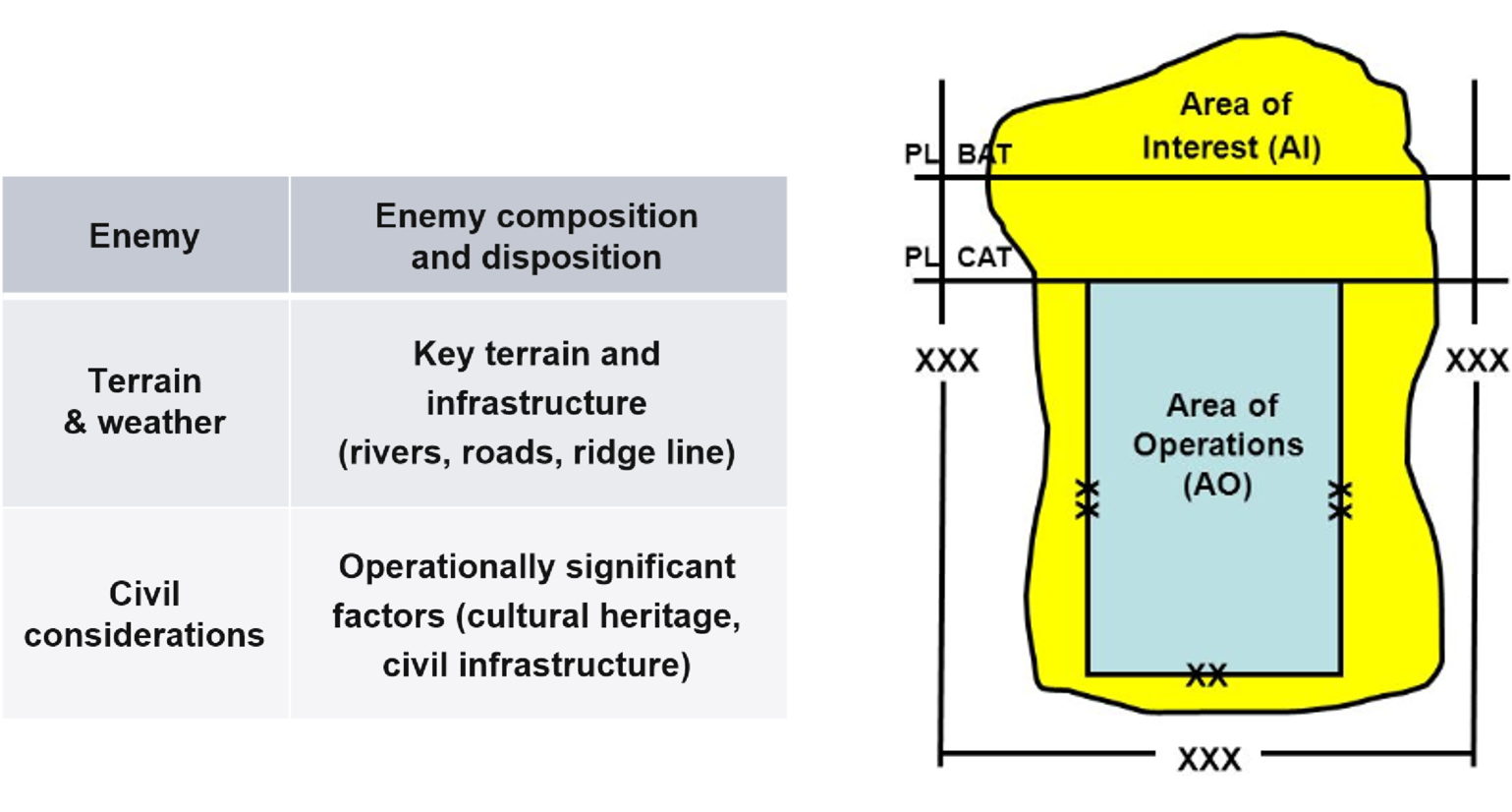}
\caption{Battlespace Evaluation for Defining the Area of Operations (AO) and Area of Interest (AI) \cite{b61}.}
\label{fig:area}
\end{figure}

\subsubsection{CoA analysis}
Fig. \ref{fig:wargame} shows that CoA analysis involves conducting a war-gaming simulation between the most likely enemy CoA and multiple friendly CoAs previously derived during the CoA development phase, in order to assess the likelihood of success of the friendly operation. Currently, this simulation is performed using rule-based M\&S models . However, such simplified models struggle to adequately reflect the vast number of variables and considerations present in the modern battlefield. Therefore, there is a growing need for AI-based war-gaming simulation models.

By applying artificial intelligence into wargame simulation process, it becomes possible to conduct simulations using AI models trained on optimal CoAs selected through tactical discussions, doctrinal principles, and historical combat data. This approach enables more accurate estimation of each CoA's strengths, weaknesses, and probability of success, thereby allowing for a more rapid comparison and analysis of multiple CoAs.

\subsubsection{CoA selection}
The final phase, CoA selection, involves comparing and analyzing multiple CoAs using an AI-based war-gaming simulation model to ultimately identify and recommend the optimal friendly CoA. As illustrated in the Fig. \ref{fig:wargame}, this process evaluates and compares CoAs based on key assessment criteria such as probability of success, estimated damage rates, and the reliability of each CoA. These strengths and weaknesses of each CoA are considered comprehensively to determine their relative merits.

At this stage, the predicted outcomes of each CoA must be presented to commanders and staff through an interpretable and explainable interface, enabling them to make informed decisions with confidence. The system should provide clear justifications for how each outcome was derived, indicate at which phases advantages or disadvantages emerged, and identify any operational aspects that may require adjustment. Such transparency is essential for the commander to assess whether a given CoA is truly optimal.

Therefore, rather than relying solely on an end-to-end approach, the application of explainable artificial intelligence (XAI) is essential. This requirement is not limited to the CoA selection phase but is a common necessity throughout the entire CoA automation system.


\subsection{IPB}
\subsubsection{Battlespace evaluation}

Battlespace Analysis focuses on identifying key characteristics of the operational environment—including enemy forces, terrain, and weather—and defining the battlespace by distinguishing between the area of operations (AO) and areas of interest (AI). This phase involves analyzing the composition, disposition, and general characteristics of enemy forces currently in confrontation.

For terrain and weather, features such as rivers, transportation networks, and elevated terrain within the AO are examined to assess their potential impact on both friendly and enemy operations. Civil considerations focus on identifying non-military factors that may influence operations, such as the presence of cultural heritage sites, hospitals, and other relevant civilian infrastructure.

\begin{figure}[t!] 
\centering
\includegraphics[width=1.0\linewidth]{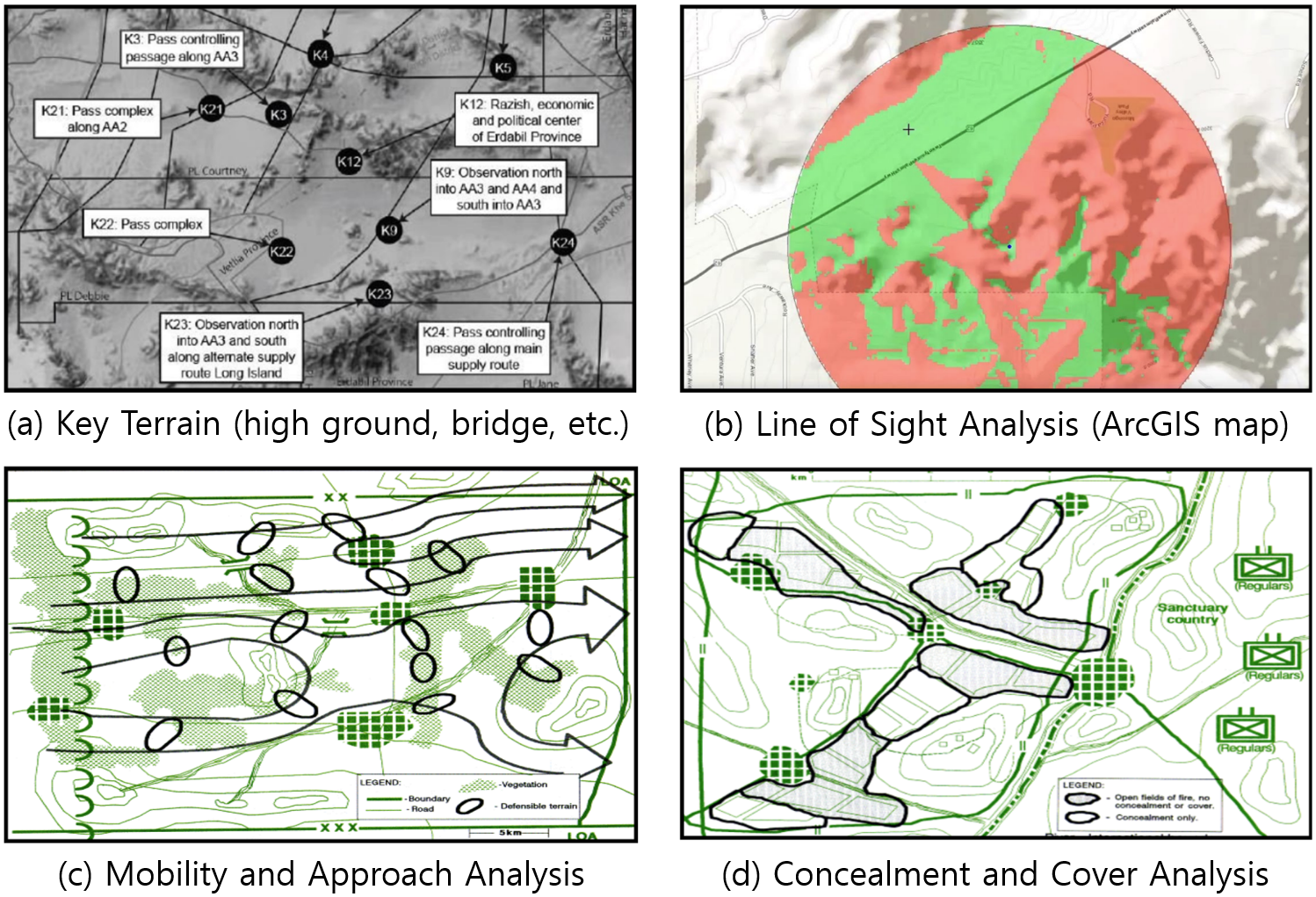}
\caption{Battlespace Analysis Based on Terrain Analysis \cite{b61}.}
\label{fig:key_terrain}
\end{figure}

It is important to note that the enemy analysis conducted in this phase does not incorporate real-time identified enemy forces. Rather, it focuses on a general assessment of the organization, disposition, and strengths and weaknesses of the opposing forces currently in confrontation. The analysis involving real-time enemy identification is conducted separately in Phase 3: enemy capabilities assessment. As shown in Figure, battlespace definition includes not only the friendly unit’s designated area of operations but also adjacent areas that, while outside the AO, may still impact the mission.

Battlespace evaluation is based not on real-time collected information, but on continuously accumulated data regarding enemy forces, terrain, and weather. Since it focuses on identifying only key features and defining the battlespace area, results can be derived using a relatively simple rule-based analysis model rather than an AI-based model.

\subsubsection{Battlespace analysis}
Battlefield analysis involves a detailed examination of terrain and weather conditions. While the previous battlespace evaluation focuses on identifying general characteristics, this phase conducts a far more comprehensive analysis. Based on the results of this in-depth assessment, a comprehensive terrain analysis map is developed.

\begin{figure}[t!] 
\centering
\includegraphics[width=0.975\linewidth]{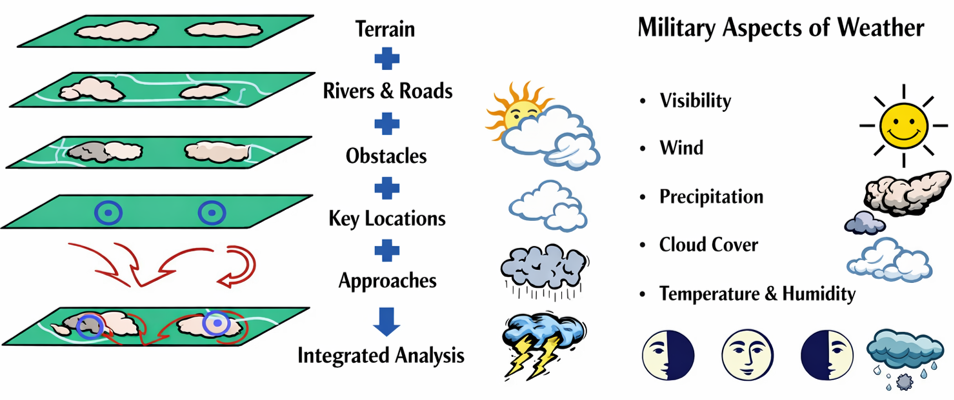}
\caption{Battlespace Analysis Using Terrain and Weather Information.}
\label{fig:terrain_weather}
\end{figure}

As shown in  Fig. \ref{fig:key_terrain}, this map is constructed by overlaying multiple analytical layers, including high ground, rivers and roads, obstacles, key terrain features, and avenues of approach. When weather analysis results are incorporated into this layered map, it becomes a complete operational area analysis map, as shown in Fig. \ref{fig:terrain_weather}.

The creation of terrain-specific layers and their integration into a composite map has already been implemented in military digital mapping systems, as well as in commercial platforms such as Google Maps and Naver Maps. Accordingly, it is technically feasible—leveraging current AI technologies—to analyze key terrain features, generate terrain analysis maps, and incorporate real-time weather data to produce an integrated operational area map.

\subsubsection{Enemy capability assessment}
The enemy capability assessment is a phase that evaluates the enemy’s operational capabilities based on the results of battlefield analysis—including assessments of enemy composition, deployment, and combat power—combined with real-time analysis of observed enemy activities. A central component of this phase is estimating the enemy’s expected disposition using doctrinal templates, as illustrated in the figure, and generating an updated enemy situation map by integrating both the actual topography of the operational area and real-time identified enemy information.

\begin{figure*}[t!] 
\centering
\includegraphics[width=0.975\linewidth]{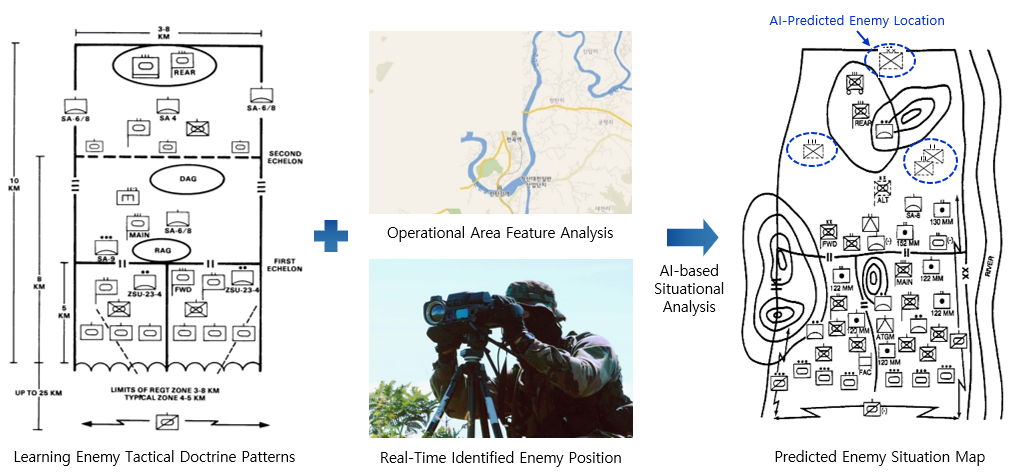}
\caption{AI-Enabled Enemy Capability Assessment in IPB Step 3 Integrating Doctrinal Patterns and Real-Time Enemy Observations.}
\label{fig:capa}
\end{figure*}

At this stage, it is important to complete the enemy situation map by anticipating the deployment of unidentified enemy units through additional consideration of the actual terrain and the positions of enemy forces identified in real time. This process can be implemented through the application of AI. Fig. \ref{fig:capa} provides an example of this approach. According to enemy doctrine, command posts are typically positioned on elevated terrain along the main axis of attack, while artillery units are located on the reverse slopes of high ground. When actual terrain data is applied to the enemy template, the initial enemy positions are first updated to reflect terrain constraints. Subsequently, if additional enemy information identified by friendly observation assets is incorporated, a generative AI model estimates the locations of unidentified enemy units, resulting in a second update to the enemy situation map.

This can be implemented as a procedure in which the enemy configuration and disposition defined by doctrine are trained into AI model. Actual terrain data and real-time identified enemy information are then used as input to generate a final enemy situation map as output. Enemy forces identified by observation assets are shown as solid lines, whereas those estimated through doctrinal and terrain analysis are shown as dashed lines.

\begin{figure}[b!] 
\centering
\includegraphics[width=1.0\linewidth]{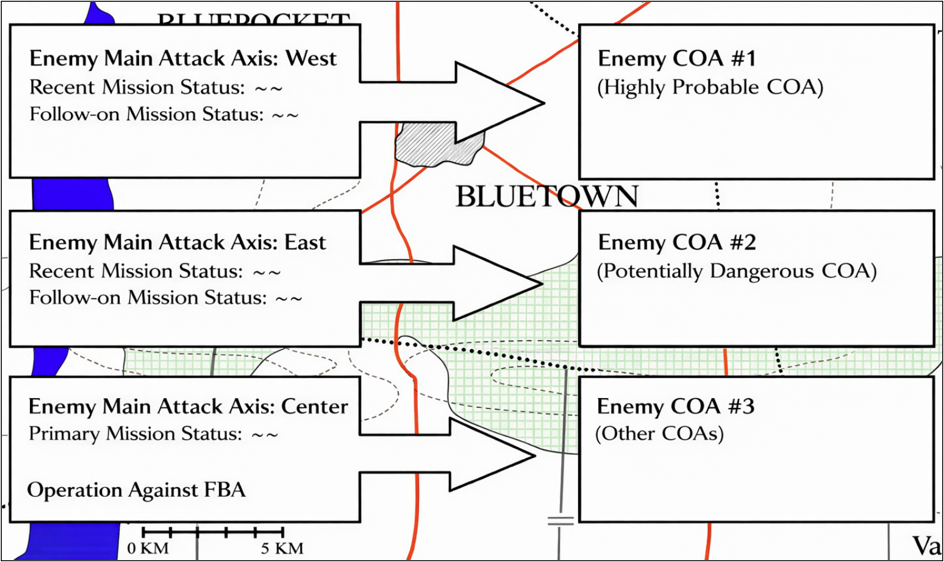}
\caption{Development of Feasible Enemy CoAs and Comparative Analysis of Strengths and Weaknesses.}
\label{fig:compare}
\end{figure}

\begin{figure*}[t!] 
\centering
\includegraphics[width=1.0\linewidth]{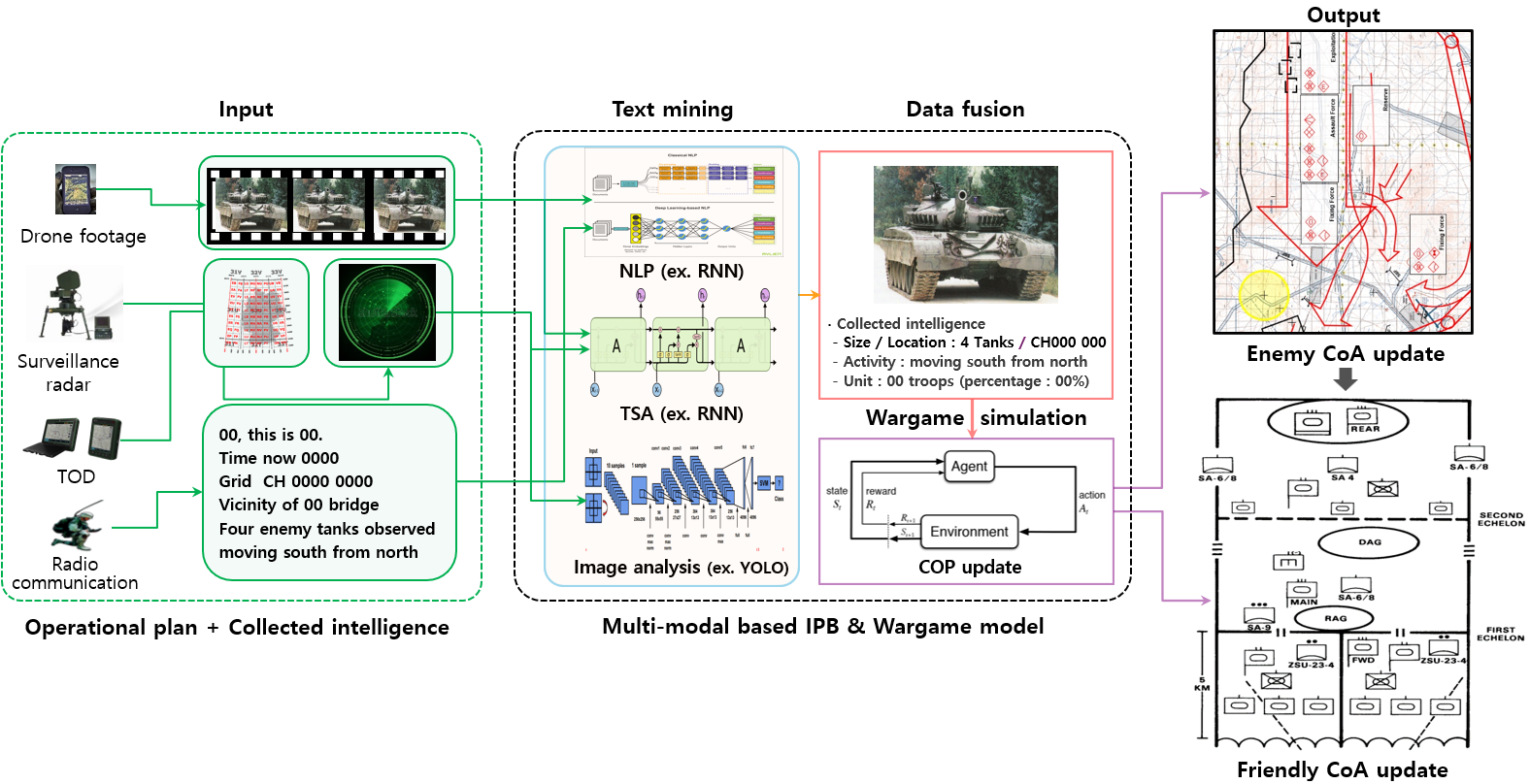}
\caption{AI-Enabled Multi-Modal IPB and Wargaming Architecture for Updating Enemy and Friendly Courses of Action.}
\label{fig:architecture}
\end{figure*}

\begin{figure*}[t!] 
\centering
\includegraphics[width=1.0\linewidth]{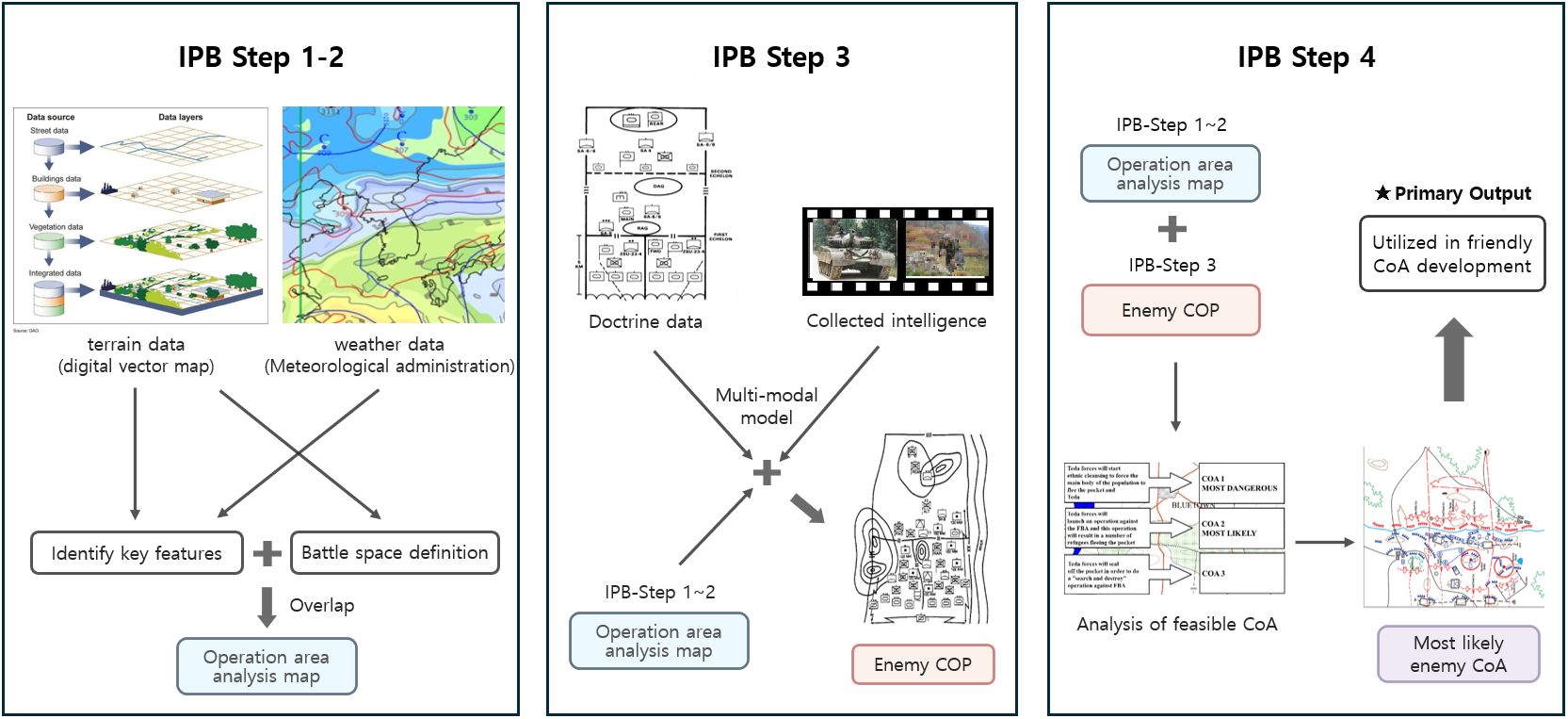}
\caption{AI-Driven IPB Workflow Across Steps 1–4 from Battlespace Analysis to Enemy CoA Estimation.}
\label{fig:ipb_app}
\end{figure*}

\subsubsection{Enemy CoA analysis}
The final phase, enemy CoA analysis, is formulating several potential enemy CoAs, comparing them, prioritizing each option, and selecting the most likely CoA as shown in the Fig. \ref{fig:compare}. To this end, a range of feasible enemy CoAs are first developed. At this stage, AI can be applied to generate multiple CoAs and evaluate each one by analyzing its likelihood of adoption and associated threat level in probabilistic terms. Since this process corresponds to Phases 3 and 4 of the planning procedure—CoA Development and CoA Analysis—it can be implemented using the same AI model, provided that the model has been trained on enemy doctrinal knowledge.

\section{Final Architecture of the Automated CoA Planning System}    

The architecture for the development of the CoA automation system proposed in this study, which corresponds to the Planning phase of the Operation Process, is illustrated in Fig. \ref{fig:architecture}. In this architecture, operational orders from higher-echelon units, along with real-time enemy information collected from various surveillance assets, serve as input to a multimodal battlefield information analysis and war-gaming model.

\begin{figure*}[t!] 
\centering
\includegraphics[width=0.995\linewidth]{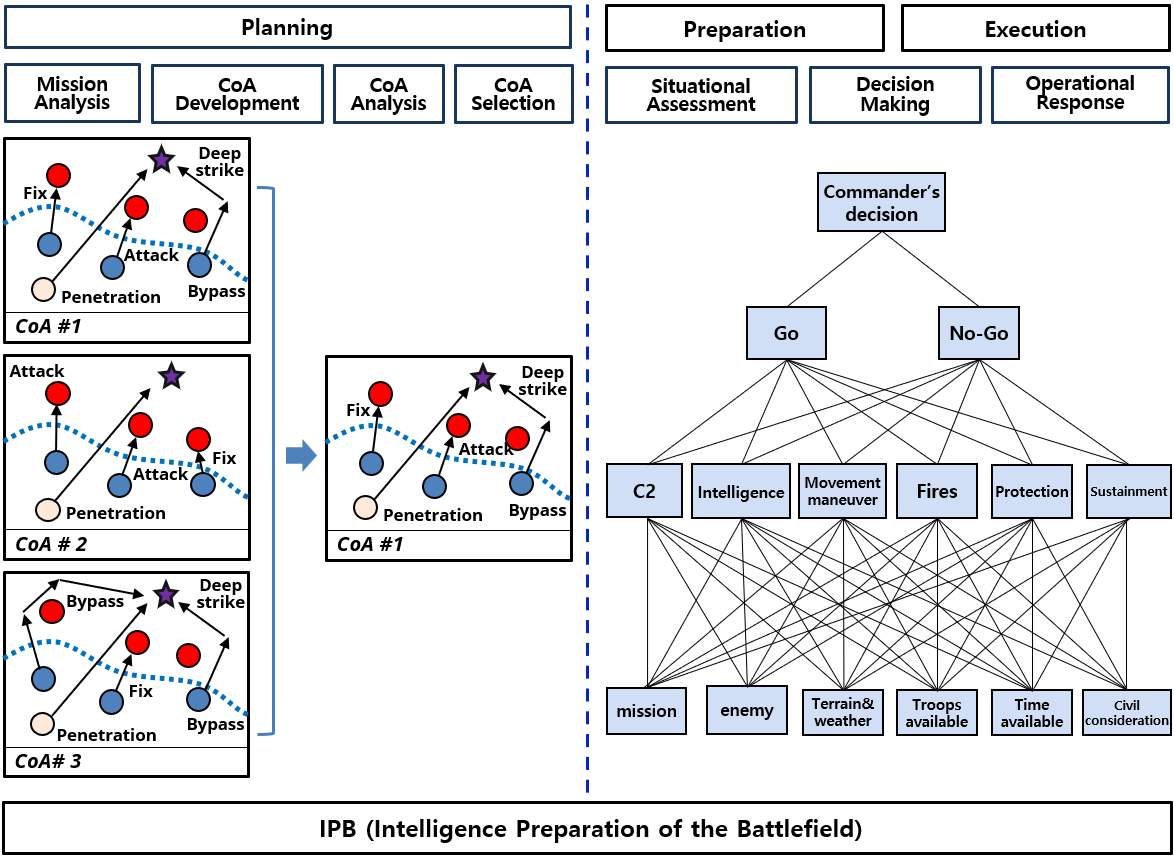}
\caption{AI-Driven IPB-Based Decision-Making Framework for Course of Action Development and Execution.}
\label{fig:final_architecture}
\end{figure*}

Specifically, image and video data are processed through computer vision models, while voice and textual data are handled by natural language processing (NLP) models. Sequential and continuously incoming time-series data are analyzed using recurrent neural networks (RNNs). The results from each of these specialized models are then integrated through data fusion techniques to update enemy information, which in turn is used to continuously generate and refine a real-time enemy situation map. Finally, based on the most recent version of this updated enemy situation map, an AI-driven war-gaming simulation is conducted to evaluate potential friendly CoA and derive the optimal strategy. This system enables intelligent, data-driven decision-making in dynamic and uncertain operational environments.

To achieve this, the development of an automated system for the IPB process—which is conducted continuously throughout the entire CoA planning phase—must be established as a prerequisite. The architecture for developing the AI-based automated IPB process is illustrated in Fig. \ref{fig:ipb_app}. In Phases 1 and 2 (Operational Environment Evaluation and Analysis), AI algorithms are applied to commercial or military digital map images to extract key features—such as roads, rivers, and mountainous terrain—layer by layer. These extracted features are then superimposed to generate a comprehensive analysis map. Additionally, meteorological data layers containing numerical information such as precipitation, temperature, and wind speed are incorporated to complete the Operational Area Analysis Map.

In Phase 3, an adversary doctrinal template is generated by training AI models on enemy doctrine data. The analysis output from Phases 1 and 2, along with real-time adversary information (e.g., location and scale), is then integrated. At this stage, advanced AI models such as Generative Adversarial Networks (GANs) are employed to synthesize a dynamic Enemy Situation Map based on fused battlefield intelligence.

Finally, in Phase 4, all battlefield information data—namely the Operational Area Analysis Map and the Enemy Situation Map—are used as input for an AI-based war-gaming simulation model. This model evaluates multiple CoAs by simulating their success probabilities and threat levels. The CoA with the highest likelihood of success is selected. Subsequently, key features extracted in Phases 1 through 3 are used to conduct a detailed assessment of the strengths and weaknesses of the selected plan.

Through comprehensive data fusion, this architecture enables the generation of a realistic and actionable enemy situation map, facilitating more accurate and intelligent decision-making in modern warfare.

Fig. \ref{fig:final_architecture} presents the final integrated architecture for the development of the CoA automation system, encompassing all phases of the operation process: planning, preparation, and execution. This figure outlines detailed sub-architectures corresponding to each phase of automation system development. As shown at the bottom of the Fig. \ref{fig:final_architecture}, the IPB process is conducted continuously across all stages of operation process. Through the development of AI-based automation, the system aims to provide real-time IPB analysis results that are significantly faster than conventional human-driven methods.

In the planning phase, depicted on the left side of the Fig. \ref{fig:final_architecture}, the AI based CoA automation system autonomously carries out mission analysis, CoA development, CoA analysis, and CoA selection. This process must account for the real-time integration of the six functions of combat power—C2, Intelligence, Movement and Maneuver, Fires, Protection, and Sustainment—as well as the METT-TC factors: Mission, Enemy, Terrain and Weather, Troops Available, Time Available, and Civil Considerations. Given the inherent uncertainty in battlefield environments and the wide range of dynamic variables and considerations involved, the development of a fully autonomous, AI-based decision-making model remains a significant challenge at present. However, it is both technically and ethically viable to develop AI systems that assist the commander’s decision-making process rather than replace it.

Therefore, current efforts should focus on developing AI systems that support, rather than replace, commanders in the decision-making process. Ultimately, by combining the proposed IPB automation and CoA generation systems with future automation solutions for the preparation and execution phases, it will be possible to realize an advanced AI-driven command decision-making system that may exceed human capabilities. Nonetheless, considering the ethical implications associated with life-and-death decisions, the final authority must always rest with human commanders.

\section{Conclusion}     
In conclusion, this study proposed an architecture for developing an AI-based CoA automation system, with a particular emphasis on a doctrine-based approach. Unlike previous studies, it offers a detailed step-by-step architecture grounded in military doctrine and introduces a novel framework that integrates these stages into a unified system. By structuring the CoA automation process in accordance with doctrinal phases, the proposed architecture enables systematic analysis and supports operationally meaningful decision-making.

In particular, this study emphasizes that AI technologies should be applied selectively and appropriately to each phase of the planning process, rather than relying on a single end-to-end model. By integrating AI-driven IPB, mission analysis, CoA development, wargaming-based CoA analysis, and CoA selection within a single framework, the proposed system preserves explainability and traceability while enhancing decision-support efficiency. This phase-aligned design allows commanders and staff to understand not only the recommended CoA but also the analytical basis behind each decision, which is essential in military operational environments.

Although the proposed architecture demonstrates the feasibility of AI-assisted CoA automation, practical challenges remain, including data availability, model validation, and robustness under uncertain operational conditions. Future work should focus on refining phase-specific AI models and validating the proposed framework through simulation-based experimentation. Nevertheless, this study provides a foundational architectural reference for the incremental development of AI-based CoA planning systems aligned with military doctrine and modern multi-domain operational requirements.


\begin{IEEEbiography}[{\includegraphics[width=1in,height=1.25in,clip,keepaspectratio]{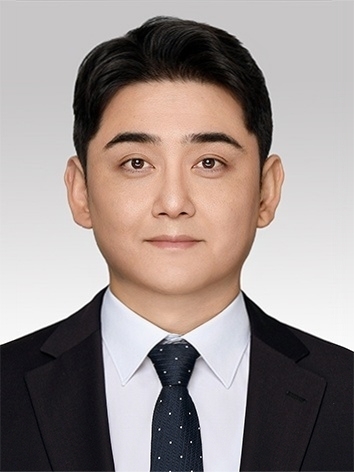}}]{Ji-il Park} received the B.S. degree in mechanical engineering from Korea Military Academy, Seoul, South Korea, in 2005, and the M.E. and Ph.D. degrees in mechanical engineering from KAIST, Daejeon, South Korea, in 2013 and 2022, respectively. He served as an Assistant Professor with the Department of Mechanical Engineering, Korea Military Academy, until 2019. He served as an Officer in charge of robot and AI at Defense Innovation Bureau, Office of the Vice Minister of National Defense, Ministry of National Defense, from 2022 to 2023, and served at the Defense AI Center in the Agency for Defense Development from 2024 to 2025, where he was responsible for artificial intelligence policy and the requirements planning of ground AI weapon systems. He is currently a Lieutenant Colonel with the Ministry of National Defense and holds a concurrent position at the Department of Smart Mobility Engineering, Inha University.
\end{IEEEbiography}

\begin{IEEEbiography}[{\includegraphics[width=1in,height=1.25in,clip,keepaspectratio]{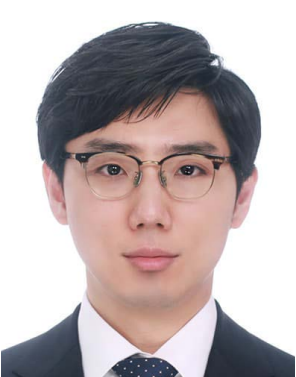}}]{Inwook Shim} received the B.S. degree in computer science from Hanyang University, in 2009, the M.S. degree in Robotics Program in electrical engineering from KAIST,
in 2011, and the Ph.D. degree in electrical engineering from the Division of Future Vehicles, KAIST, in 2017. From 2017 to 2022, he worked as a Research Scientist at Agency for Defense Development, South Korea. He is currently an Assistant
Professor with the Department of Smart Mobility Engineering, Inha University, Incheon, South Korea. His research interests include 3D vision for autonomous systems and deep learning. He was a member of Team KAIST, which took first place at the DARPA Robotics Challenge Finals, in 2015. He was a recipient of the Qualcomm Innovation Award, and a NI finalist at the NI Student Design Showcase. He received the KAIST Achievement Award of Robotics, and the Creativity and Challenge Award from KAIST.
\end{IEEEbiography}

\begin{IEEEbiography}[{\includegraphics[width=1in,height=1.25in,clip,keepaspectratio]{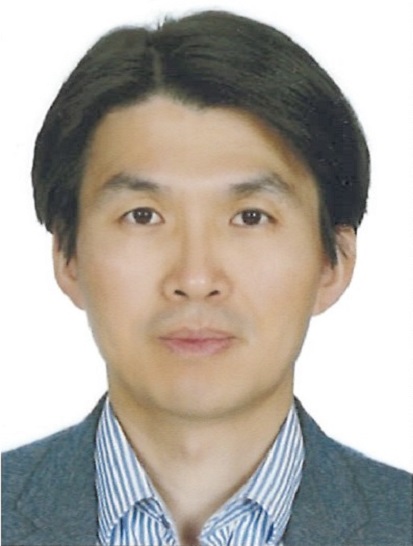}}]{Chong Hui Kim} received a B.S., M.S., and Ph.D. degrees in electrical engineering from Korea Advanced Institute of Science and Technology(KAIST), Daejeon, South Korea, in 1999, 2001, and 2007. He worked as a Postdoctoral Researcher with the Human-Robot Interaction Research Center at KAIST in 2007. He is currently a principal researcher at Agency for Defense Development(ADD) and is in charge of chief of 1st Division at the Defense AI R\&D Institute at ADD, Daejeon, South Korea. His research interests include robotics, autonomous navigation, system architecture, and artificial intelligence of defense systems.
\end{IEEEbiography}



\begin{thebibliography}{00}

 
\bibitem{b1}
U.S.~Department~of~Defense,
\textit{Joint Publication 5-0: Joint Planning},
Washington, DC, USA, Dec. 2020.

\bibitem{b2}
NATO Allied Command Transformation,
\textit{Artificial Intelligence in Multi-Domain Operations},
NATO ACT Report, Norfolk, VA, USA, 2021.

\bibitem{b3}
P.~Scharre,
\textit{Army of None: Autonomous Weapons and the Future of War},
W.~W.~Norton \& Company, New York, NY, USA, 2018.

\bibitem{b4}
H.-X.~Li,
``Exploration of wargaming and AI applications in military decision-making,''
in \textit{Proc. 2025 International Conference on Military Technologies (ICMT)},
May 2025, doi: 10.1109/ICMT65201.2025.11061360.

\bibitem{b5}
U.S.~Army Futures Command,
\textit{Operationalizing AI for Decision Dominance: AI Strategy Implementation Plan},
AFC Technical Report No.~22-103, Austin, TX, USA, 2022.

\bibitem{b6}
U.S.~Army~Futures~Command,
\textit{Future Study Plan 2019: Operationalizing Artificial Intelligence for Multi-Domain Operations},
Futures and Concepts Center, Fort Eustis, VA, USA, Aug. 2019.

\bibitem{b7}
Ministry of National Defense, Republic of Korea,
\textit{Defense Innovation 4.0 Brochure},
[Online]. Available:
\url{https://www.mnd.go.kr/mbshome/mbs/mndEN/download/Defense_Innovation_4.0_brochure_EN.pdf}
 
\bibitem{b8}
P.~J.~Schwartz, D.~V.~O'Neill, M.~E.~Bentz, A.~Brown, B.~S.~Doyle,
O.~C.~Liepa, R.~Lawrence, and R.~D.~Hull,
``AI-enabled wargaming in the military decision making process,''
in \textit{Artificial Intelligence and Machine Learning for Multi-Domain Operations Applications II},
vol.~11413, pp.~118--134, SPIE, Apr. 2020.
 
\bibitem{b9}
L.~Groud and A.~Kott,
``Course of Action Display and Evaluation Tool (CADET) enhancements,''
in \textit{Proc. Winter Simulation Conference},
Orlando, FL, USA, Dec. 2000, pp.~123--130.
 


\bibitem{b35} M.~Ben-Bassat and A.~Freedy, “Knowledge requirements and management in expert decision support systems for (military) situation assessment,” \textit{IEEE Transactions on Systems, Man, and Cybernetics}, vol.~12, no.~4, pp.~479--490, 1982.

\bibitem{b36} S.~H.~Liao, “Case-based decision support system: Architecture for simulating military command and control,” \textit{European Journal of Operational Research}, vol.~123, no.~3, pp.~558--567, 2000.

\bibitem{b37} F.~Min, P.~Ma, and M.~Yang, “A knowledge-based method for the validation of military simulation,” in \textit{Proc. 2007 Winter Simulation Conf.}, pp.~1395--1402, Dec. 2007.

\bibitem{b38} I.~Cil and M.~Mala, “MABSIM: A multi agent based simulation model of military unit combat,” in \textit{Proc. 2009 Second International Conf. on the Applications of Digital Information and Web Technologies}, pp.~731--736, Aug. 2009.

\bibitem{b39} V.~Middleton, “Simulating small unit military operations with agent-based models of complex adaptive systems,” in \textit{Proc. 2010 Winter Simulation Conf.}, pp.~119--134, Dec. 2010.

\bibitem{b40} G.~Secinti, P.~B.~Darian, B.~Canberk, and K.~R.~Chowdhury, “Resilient end-to-end connectivity for software defined unmanned aerial vehicular networks,” in \textit{Proc. 2017 IEEE 28th Annual International Symposium on Personal, Indoor, and Mobile Radio Communications (PIMRC)}, pp.~1--5, Oct. 2017.

\bibitem{b41} M.~Doniec, M.~Angermann, and D.~Rus, “An end-to-end signal strength model for underwater optical communications,” \textit{IEEE Journal of Oceanic Engineering}, vol.~38, no.~4, pp.~743--757, 2013.

\bibitem{b42} L.~Baranyai, E.~G.~Cuevas, S.~Davidow, C.~Demaree, and P.~DiCaprio, “End-to-end network modeling and simulation of integrated terrestrial, airborne and space environments,” in \textit{Proc. 2005 IEEE Aerospace Conf.}, pp.~1348--1353, Mar. 2005.

\bibitem{b43} K.~D.~Tuchs, T.~Halmai, and M.vanSelm, “Multi-security domain management integration architecture for end-to-end service management in military networks,” in \textit{Proc. 2011-MILCOM 2011 Military Communications Conf.}, pp.~1375--1380, Nov. 2011.

\bibitem{b44} F.~E.~Morgan, B.~Boudreaux, A.~J.~Lohn, M.~Ashby, C.~Curriden, K.~Klima, and D.~Grossman, \textit{Military applications of artificial intelligence}, RAND Corporation, Santa Monica, 2020.

\bibitem{b45} Y.~Zhang, Z.~Dai, L.~Zhang, Z.~Wang, L.~Chen, and Y.~Zhou, “Application of artificial intelligence in military: from projects view,” in \textit{Proc. 2020 6th International Conf. on Big Data and Information Analytics (BigDIA)}, pp.~113--116, Dec. 2020.

\bibitem{b46} A.~B.~Rashid, A.~K.Kausik, A.AlHassanSunny, and M.~H.~Bappy, “Artificial intelligence in the military: An overview of the capabilities, applications, and challenges,” \textit{International Journal of Intelligent Systems}, vol.~2023, no.~1, p.~8676366, 2023.

\bibitem{b47} W.~Wang, H.~Liu, W.~Lin, Y.~Chen, and J.~A.~Yang, “Investigation on works and military applications of artificial intelligence,” \textit{IEEE Access}, vol.~8, pp.~131614--131625, 2020.


\bibitem{b47a} P.~J.~Schwartz, D.~V.~O'Neill, M.~E.~Bentz, A.~Brown, B.~S.~Doyle, Q.~C.~Liepa, and R.~D.~Hull, “AI-enabled wargaming in the military decision making process,” in \textit{Artificial Intelligence and Machine Learning for Multi-Domain Operations Applications II}, vol.~11413, pp.~118--134, Apr. 2020.

\bibitem{b47b} V.~G.~Goecks and N.~Waytowich, “CoA-GPT: Generative pre-trained transformers for accelerated course of action development in military operations,” in \textit{Proc. 2024 International Conference on Military Communication and Information Systems (ICMCIS)}, IEEE, pp.~1--10, Apr. 2024.

\bibitem{b47c} K.~Payne, “AI Arms and Influence: Frontier Models Exhibit Sophisticated Reasoning in Simulated Nuclear Crises,” \textit{arXiv preprint arXiv:2602.14740}, 2026.

\bibitem{b47d} T.~Copp, E.~Dwoskin, and I.~Duncan, “Anthropic’s AI tool Claude central to U.S. campaign in Iran, amid a bitter feud,” \textit{The Washington Post}, Mar. 4, 2026.

\bibitem{b48} M.~C.~Horowitz, L.~Kahn, and C.~Mahoney, “The future of military applications of artificial intelligence: A role for confidence-building measures?” \textit{Orbis}, vol.~64, no.~4, pp.~528--543, 2020.


\bibitem{b50} B.~Hallaq, T.~Somer, A.~M.~Osula, K.~Ngo, and T.~Mitchener-Nissen, “Artificial intelligence within the military domain and cyber warfare,” in \textit{Proc. Eur. Conf. Inf. Warf. Secur. (ECCWS)}, pp.~153--157, Jun. 2017.


 

\bibitem{b55}
S.~Yoon,
\textit{3D Combined Flight Path Algorithm for UAV Considering Turning and Elevation},
M.S. thesis, Dept. of Mechanical Engineering, TU Korea, Siheung, South Korea, 2019.


\bibitem{b59}
U.S.~Army,
\textit{ADP 5-0: The Operations Process},
Washington, DC, USA, Jul. 2019.

\bibitem{b60}
U.S.~Department~of~the~Army,
\textit{ATP 2-01.3: Intelligence Preparation of the Battlefield},
Headquarters, Department of the Army, Washington, DC, USA, Mar. 2019.

\bibitem{b61}
U.S.~Department~of~the~Army,
\textit{FM 34-130: Intelligence Preparation of the Battlefield},
Headquarters, Department of the Army, Washington, DC, USA, Jul. 1994.



\end{thebibliography}
\end{document}